\documentclass[a4paper]{article}

\usepackage[pages=all, color=black, position={current page.south}, placement=bottom, scale=1, opacity=1, vshift=5mm]{background}
\usepackage{xcolor}
\pagecolor{white}
\usepackage[margin=1in]{geometry} 

\usepackage{amsmath}
\usepackage{amsthm}
\usepackage{amssymb}
\usepackage{cite} 
\usepackage[utf8]{inputenc}
\usepackage{hyperref}
\hypersetup{
	unicode,
	pdfauthor={Author One, Author Two},
	pdftitle={Anomaly Detection in OKTA Logs using Autoencoders},
	pdfkeywords={autoencoder, deep learning, cybersecurity, anomaly detection, okta},
	pdfproducer={LaTeX},
	pdfcreator={pdflatex}
}

\usepackage{graphicx, color}
\graphicspath{{./images/}}

\usepackage{algorithm, algpseudocode} 
\usepackage{mathrsfs} 

\usepackage{lipsum}

\usepackage{listings}
\DeclareFixedFont{\ttb}{T1}{txtt}{bx}{n}{12} 
\DeclareFixedFont{\ttm}{T1}{txtt}{m}{n}{12}  

\usepackage{color}
\definecolor{deepblue}{rgb}{0,0,0.5}
\definecolor{deepred}{rgb}{0.6,0,0}
\definecolor{deepgreen}{rgb}{0,0.5,0}

\lstdefinestyle{pythonstyling}{
language=Python,
basicstyle=\tiny,
morekeywords={self},              
keywordstyle=\tiny\color{deepblue},
emph={MyClass,__init__},          
emphstyle=\tiny\color{deepred},    
stringstyle=\color{deepgreen},
frame=tb,                         
showstringspaces=false,
breaklines=true
}

\theoremstyle{thmstyleone}%
\theoremstyle{thmstyletwo}%
\theoremstyle{thmstylethree}%

\title{Anomaly Detection in OKTA Logs using AutoEncoders}
\author{Jericho E. Cain PhD$^1$\and Hayden Beadles$^2$\and Karthik Venkatesan$^3$}

\date{
	$^1$Adobe \\ \texttt{jerichoc@adobe.com}\\%
	$^2$Adobe \\ \texttt{beadles@adobe.com}\\%
	$^3$Adobe \\ \texttt{kavenkat@adobe.com}\\[2ex]%
}

\begin{document}


\maketitle



\begin{abstract}Okta logs are used today to detect cybersecurity events using various rule-based models with restricted look back periods. These functions have limitations, such as a limited retrospective analysis, a predefined rule set, and susceptibility to generating false positives. To address this, we adopt unsupervised techniques, specifically employing autoencoders. To properly use an autoencoder, we need to transform and simplify the complexity of the log data we receive from our users. This transformed and filtered data is then fed into the autoencoder, and the output is evaluated.
\end{abstract}

\section{Introduction}
Okta's Behavior Detection function, which is a current offering in Okta's commercial SSO product, operates on a rules-based engine that detects potential cybersecurity events by analyzing user behavior patterns within the Okta system. While this tool can be effective at identifying certain types of anomalies, it also has several limitations that can impact its overall efficacy.

One limitation to this built-in tool is that it is based on a predefined set of data and rules that may not capture all potential threat scenarios. Consequently, the tool could overlook threats that fall outside these predefined data and rules or that occur in a way that fail to trigger alerts. Additionally, as new types of threats emerge, the rules may need to be updated to effectively detect them, potentially leading to delayed detection until the rules are updated

Another limitation is the restricted look-back window. The tool is designed to analyze user behavior within a limited time frame, typically an average of 20 authentication attempts. This means that threats that occur outside of this window will not be detected.  Even more importantly, establishing a realistic \emph{baseline} that reflects a user's actual behavior is limited to this time frame alone.

The tool may also produce false positives. False positives can lead to unnecessary alerts and can create additional work for security teams. With a restricted look back window, typical behavior may be truncated and identified as an anomaly. By expanding this window to look back through the user's entire history, this typical behavior would be captured.

The Okta's Behavior Detection tool can be a useful capability of a larger cybersecurity strategy, but it is important to be aware of its limitations and to use it in conjunction with other security measures to provide comprehensive threat detection and response capabilities.

On the other hand, when dealing with large datasets, heuristic models face scalability challenges. Identifying and filtering anomalous events can reduce data volume considerably, leading to a less complex heuristic model that's easier to manage. Such a streamlined process improves scalability. However, moving towards an efficient and scalable anomalous data detector might lead us to supervised methods, which necessitate labeled data. Relying on heuristics for labeling could reintroduce the same scalability and maintenance concerns, limiting the historical data available for training.

Our goal is to detect user behavior anomalies using the event hour, day of the week, event outcome, location and application fields of the Okta System Log dataset and do so using Autoencoders. An autoencoder can be defined as a neural network that is trained to try to emulate its input in its output. This provides a useful mechanism to detect behavior that differs significantly from a specific users’ typical login behavior. We define anomalies broadly, but offer several examples to guide our approach: 

	\begin{itemize}
		\item User $x$ Logs into a meaningfully different location from normal
		\item User $x$'s behavior (for example, applications accessed) has changed meaningfully from their typical behavior, based on some established baselines.
		\item User $x$ is engaging in behavior that is specifically strange (lots of MFA requests, login failures, lockouts, etc) within a smaller window of time.
	\end{itemize}
	It is important to note that we use the term "meaningful" intentionally. To accurately detect changes in user behavior, we must first establish a baseline understanding of their typical patterns within the Okta data. This allows us to identify meaningful changes and ensure our autoencoder model accurately represents the users' behavior.  
	
\section{OKTA Data}
Our analysis of user behavior relies on the Okta System Log \cite{okta_sys_log}, a time series dataset that captures event-level log data from Okta. This data is consistently transmitted in raw JSON format to our Security Data Lake.

The Okta System Log is a complex resource to navigate. The first part to discuss is the 800+ unique event types \cite{brodksy}. Most event types are internal Okta messages used for communication within the system, but our focus lies on a subset of events that relate to user logins and users' interaction within the system.  

The events that capture a user's login and interaction can vary significantly depending on how an organization has configured their Okta environment. For instance, an organization may have a customized sign-on policy with specific rules and requirements a user must follow, such as multi-factor authentication (MFA), password protocols, or integration with an identity provider (IDP) \cite{okta_signon}. Moreover, the session attached to a user's login may have an expiration time unique to the organization  \cite{okta_config}.

Further still, certain events are specific to a particular type of behavior. For instance, the event \emph{user.account.lock} signifies when a user is locked out of the Okta system due to too many failed password login attempts. Security teams often search for known event types like these to detect anomalous behavior. As another example, a team may search the \emph{displayMessage} feature to find users that have exceeded Okta’s login rate limit  \cite{okta_useful} .

All these examples have a major gap though. They’re only looking for events within specific windows of time, without including the context of the user's history, their day-to-day login habits, or the applications and locations they may login from. This behavioral context is crucial to understanding the difference between a true anomaly and a false positive.  

With all these complexities in mind, we proceed with the first question, how, given these above constraints, can we define an initial point of entry for our user? This entry point is important not only for gaining intelligence about our users but also for testing and validating the results of our autoencoders analysis. 

\subsection{Defining a Point of Entry}
	
	We define \emph{Point of Entry} as follows:
	
	\begin{itemize}
		\item \textbf{Point of Entry}: User $x$ logs into Okta at point of time $i$
	\end{itemize}
	
	A user's total \emph{sessions} are simply the sum of the individual login sessions, grouped over a window of time $t$. We can represent that simply as follows:
	
	$$ X = med(\forall x_i)$$
	
\begin{figure}[H]
		\centering
		\includegraphics[scale=0.3]{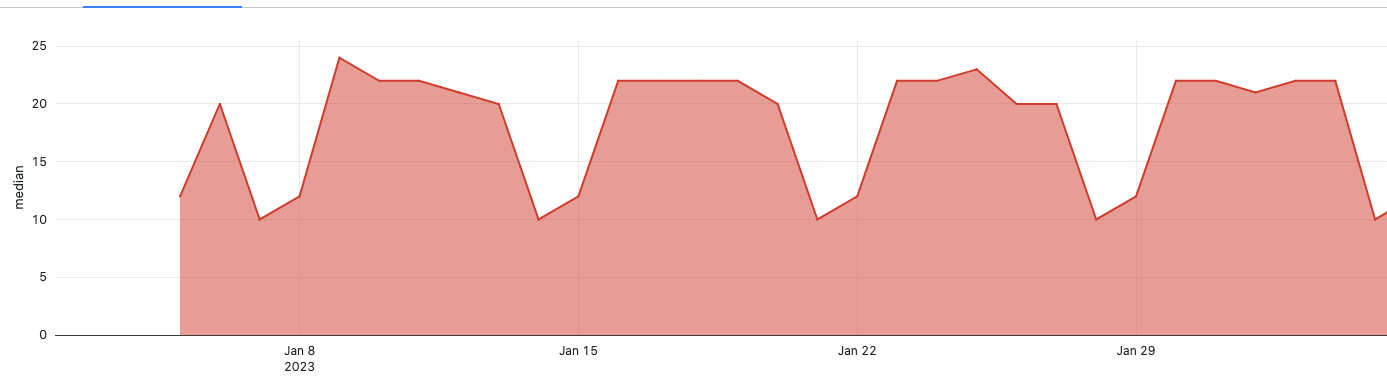}
		\caption{\label{fig:median_event_types} This plot shows the median login counts for actors by day across a sample of logins in Jan 2023. Notice the dips that occur on the weekends and the general, predictable ebb and flow of the counts.}
\end{figure}

	We take the median of these total sessions over all our users, to get an idea of the typical number of times a user logs into our system. We can then plot that over time as shown in Fig.\ref{fig:median_event_types}
	
	Envisioning a user entering our login system is analogous to an airport security system. Users login (arrive) at the Corporate SSO boundary (in this case, okta) and login to begin a session and access internal applications and tools. This arrival is predictable, and can actually be simulated via a Poisson and Exponential Distribution.
	We observe the logins appear to follow a natural pattern over time. During the weekdays, a user logs in around roughly 20 times, with dips on the weekend. 
	
	There are two events that we use to define these initial user logins, those are:
	\begin{itemize}
		\item \textbf{policy.evaluate\_sign\_on}: This event is specific to the corporate policy specified by an organization. In our case, Adobe. This event also contains critical information related to \emph{context}, which we'll also detail later 
		\item \textbf{user.authentication.sso}: User log in event via SSO. This event fires and details the result of that sso auth event (whether it succeeds or fails)
	\end{itemize}
	
	Both of these events fire every time a user logs into Okta, and thus provide our entry point for a login. They also contain critical information about the login, such as:
	\begin{itemize}
		\item \textbf{location}: This provides information on location (country, state, zip, lat / lon)
		\item \textbf{ip address}: There are a few fields that cover this, we consider client\_ip\_address, but ip address is another layer of discussion due to its high cardinality, so we won't cover it in this paper.
		\item \textbf{application}: information about the application a user may be trying to access
		\item \textbf{event}:  event context information (success / failure, etc)
		\item \textbf{device}: Device ID and information about the login in device itself.  
		\item \textbf{debug}: Contains useful information about a login from Okta itself (whether Okta suspects the event as a threat, for example). The behaviors heuristic function described in the abstract, also appears here.
	\end{itemize}
	
	We utilize this background in our approach to preparing our data for encoding
	
	\subsection{Data Engineering - Overview}
	To prepare our data for autoencoding, several essential steps are required.While we have discussed some of the complexities of the data earlier, we highlight the following steps that we undertake to ensure that our data is fully prepared:

	\begin{itemize}
		\item Events exert high variance, so we restrict our dataset to a subset of the overall Okta events, in order to establish a baseline of behavior from the noise. This is the \emph{policy.evaluate\_sign\_on} event detailed above. This subset represents a sample of the actual user behavior, and so we must accommodate that when building our features. 
		\item We utilize geohashing to encode a user’s location to a wider bounded box, to account for small variances in location due to things like ISP, VPN or general network variability in location. This allows us to assign a set of geographical regions to a user, and work with regions instead of points on a map. 
		\item We utilize frequency analysis to prepare a new variable, named known\_app or known application, to flag an application login as anomalous or not. 
		\item We encode several choice categorical variables using the String Indexer ML feature in Spark. 
		\item We use bootstrap sampling against our users to get a representation of our users that reflects the overall population. Simply choosing events within a sampled window of time is not enough to fully represent a user’s behavior. We instead sample against our user's entire login history to build a training set for our model.  
		\item We create validation sets against each of our users and introduce anomalies for location or other features of interest. We then use that to evaluate our model’s effectiveness in finding them. 
	\end{itemize} 
	
	\subsection{Data Engineering - Geohashing}
	
\begin{figure}[H]
		\centering
		\includegraphics[scale=0.6]{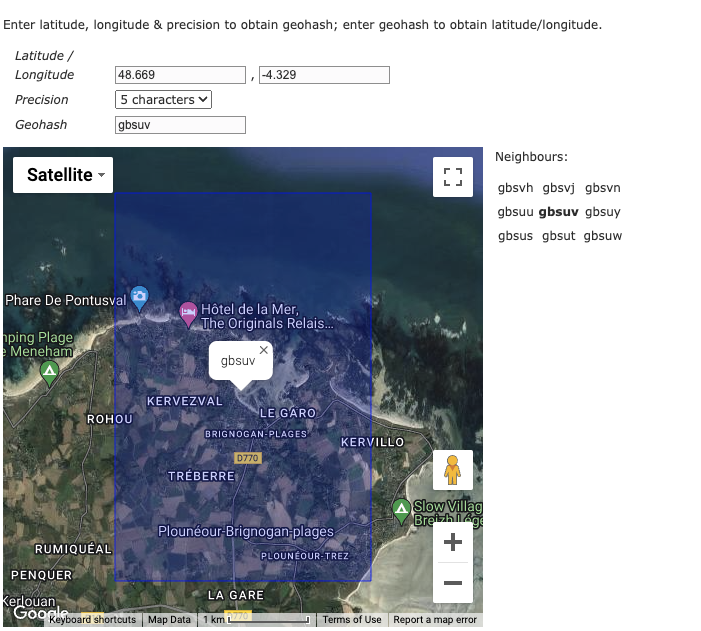}
		\caption{\label{fig:geohash_ex}This references a website: https://www.movable-type.co.uk/scripts/geohash.html. This site gives a good introduction to geohashing and a good visual introduction to precision and formulas, background on hashing algorithms, etc.}
\end{figure}

Geohashing is a technique using space filling curves such as the Z-order. The library we utilize \textbf{“python-geohash”}, employs a Z-order curve to encode a set of latitude and longitude coordinates to their respective geohashes \cite{wiki_geo}\cite{ibm_geo}. Geohashes are then represented by bounded boxes on a grid, as shown in  Fig.~\ref{fig:geohash_ex}.

Depending on the size of the hash, we can effectively control the size of the bounding box we want to set for a user's location \cite{wiki_geo}. This hash can be widened to account for a larger area based on the number of bits of precision we provide. In the example below, we set a precision of five bits, which creates an area around the latitude and longitude that is roughly 5km by 5km \cite{mov_type} . 
	
The benefit of this is quickly obvious from observation of the Okta System Log data and the variability that can arise from location. When dealing with network layers and the given complexities and constraints around translating a user's location accurately, we can safely assume that, for any given user of the system, there will be major variation in terms of the latitude and longitude coordinates, zip code, street, city, etc. However, the user still \emph{lives in that general area}. The additional noise confuses the baseline we are trying to achieve. 
	
What we want to capture is the general area of the user, \emph{the space where they reside and access our systems most of the time}. We are not as concerned with how precise it is, and a bound box can work perfectly well for our purposes. We can then show, through a basic frequency analysis, what location a user is most likely at. A geohash represents a fast and accurate way to represent this.  
	
\begin{figure}[H]
		\centering
		\includegraphics[scale=0.6]{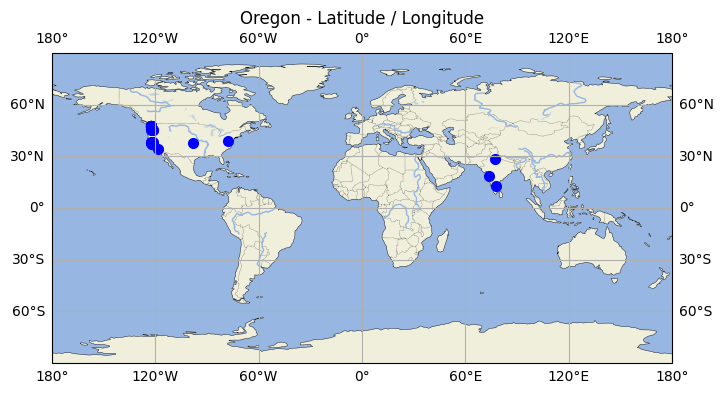}
		\caption{\label{fig:lat_lon_img}This map shows a set of logins, referenced as blue points, for a single actor. This actor's location is known as Oregon, but it can't be easily seen, due to the noise of the login data. }
\end{figure}

\begin{figure}[H]
		\centering
		\includegraphics[scale=0.6]{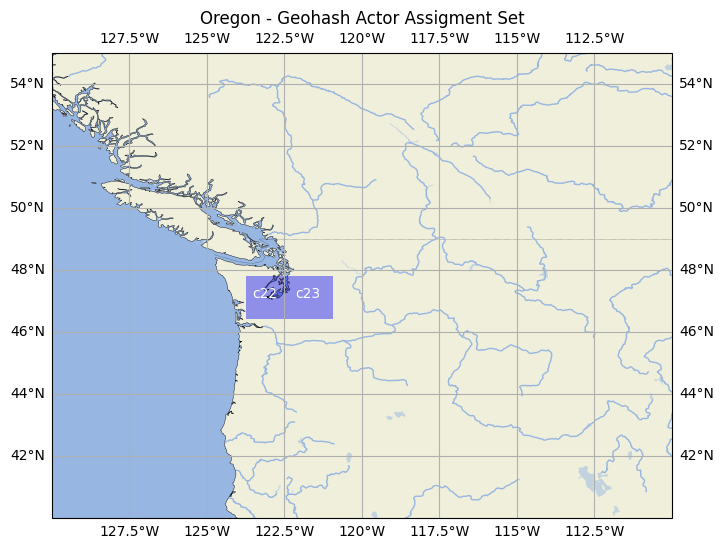}
		\caption{\label{fig:geohash_visual_after}After changing lat / lon locations to geohashes and performing some simply frequency analysis on logins per actor, we choose the login frequencies that are most relevant for an actor and adjust for the population proportion using the Wilson Score Confidence Interval. }
\end{figure}

	We can showcase this with a basic demonstration. Fig.~\ref{fig:lat_lon_img} visually shows the logins for \emph{one user} based on lat / lon coordinates. There is a major amount of variation due to all the network changes that we noted earlier.  We note that for this one user, the locations are all over the place, we see no consistent login because the lat / lon values are too specific. However, once we apply our hash this improves significantly (Note: we use a 3bit hash, roughly a 160km bounding box) as shown in Fig.~\ref{fig:geohash_visual_after}.  We see our geohash application shows much better results. We can see that one hash captures the majority of the variance in location (Oregon, U.S.), which means that using the hash establishes a much stronger baseline of where the user actually resides.

	\subsection{Data Engineering - Application}
	
	All Adobe Employees access a set of applications when utilizing Okta. Okta acts as the SSO gateway that lives in front of these apps. These apps range from internal Adobe systems, like Wiki or Jira, to intermediary or proxy systems, such as a VPN proxy or tunnel, to even very mundane or basic apps, like email or Microsoft Word \cite{okta_sso}. There are many correlated features to the set of apps a user will use, such as the employee’s job title and position, their manager, department, and the devices they are currently using. The applications a user accesses are key to understanding actor behavior, especially how it relates to potential future intrusions.  
	
	As an example of this, consider the recent Okta breach by the \emph{LAPSU\$} extortionist group back in 2020 \cite{hacker}. The group was able to gain access to Okta by exfiltrating a support engineer’s credentials, and then utilize those to infiltrate Okta’s internal systems (attempting escalation of privileges, downloading source code from Git, etc). These actions, and the apps associated with them, would represent a deviation of typical behavior from an expected support engineer’s application baseline.   
	
	We tackle encoding applications with the following steps: 
	
	Given a set of apps $X$ and users $U$
	
	\begin{equation} 
	X = {x_1, x_2 ... x_n}, U = {u_1, u_2, u_3 ... u_n} 
	\label{eq:X}
	\end{equation}
	
	The frequency distribution is then the number of times a user successfully logged into a subset of applications: 
	
	\begin{itemize}
		\item Extract all apps a user utilizes in the Okta System Log \emph{‘policy.evaluate\_sign\_on`} event and `user.authentication.sso` event 
		\item Compute the Wilson score interval around the frequency of successful logins to applications, per actor.  
		\item Create a confidence interval around this frequency for all apps and actors 
		\item From this, create a superset of applications associated with each actor, as a lookup table 
		\item Create a new feature named known\_app, that checks current daily logins with this superset of applications. Return a 1 if an app is found that is not in the superset, or a 0 otherwise.  
	\end{itemize}
	
	We create a frequency distribution of these values. We define a frequency distribution as follows.  
	
	\begin{equation} f(x) = \frac{1}{L_i - L_n} \text{for} x \subset (L_i, L_n)
	\label{eq: fx}
	\end{equation}
	
	It is the distribution of the number of times a user successfully logged into an app, divided by the total number of logins by that user 
	
\begin{figure}[H]
		\centering
		\includegraphics[scale=0.5]{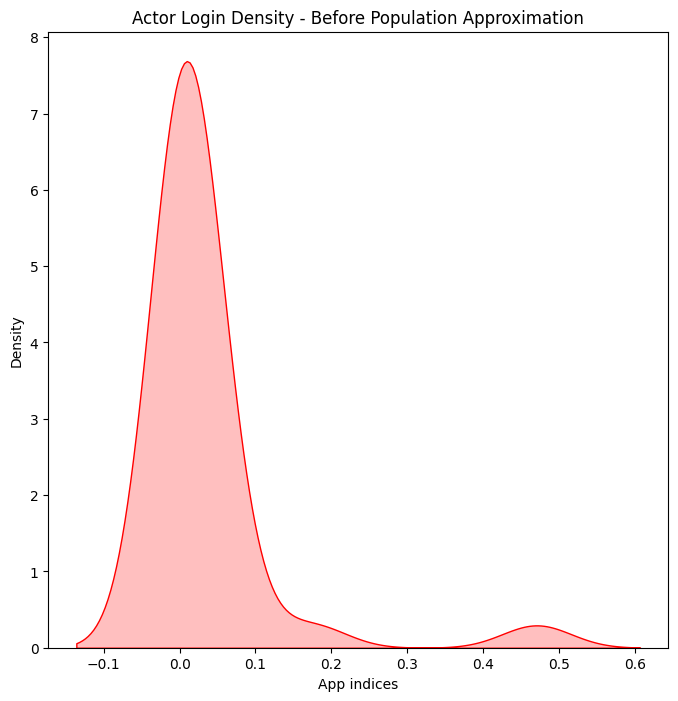}
		\caption{\label{fig:app_logins_before}This plot shows the application login density for a single actor. The x-axis reflects the encoded indices of the applications accessed, and the y-axis reflects the density, or distribution of those logins. Before estimating the effects of the population, we notice noise, especially on the right tail of the above distribution. Our goal is to get a better view of the actual application login distribution per actor, to catch anomalies. Removing the noise is therefore, our goal.}
\end{figure}

	As an example of the result, we can see the following, we divide the app logins \emph{app\_instance\_counts} by the total \emph{actor\_sum}, to get the \emph{actor\_freq} variable. We can then visualize that using a density estimation of the frequencies as shown in Fig.~\ref{fig:app_logins_before}.  This works great in theory but relates poorly to the actual set of applications and logins a user uses in the population. This is because we are only capturing a frequency distribution for applications within a sample of data (the Okta System Log), where that behavior is only a sample representation of the actor’s actual login behavior and application use.  
	
	Put another way, Okta cannot capture everything about an Adobe employee’s complete application history. Users can access other SSO tools at various times or on different devices. Thus, we need to adjust our frequency estimation with the population behavior in mind to weigh application usage correctly \cite{bookdown}. We do this by using the \emph{Wilson Score Interval}. 
	
	The Wilson Score Interval takes a probability estimate from a sample and attempts to create a confidence interval of that sample around the population using the binomial distribution \cite{wallis}. The binomial distribution is simply the probability of obtaining $k$ successes over $n$ trials. In our case, we are looking for the probability of a user successfully logging into an application $k$ times over $n$ logins. 
	
\begin{figure}[H]
		\centering
		\includegraphics[scale=0.5]{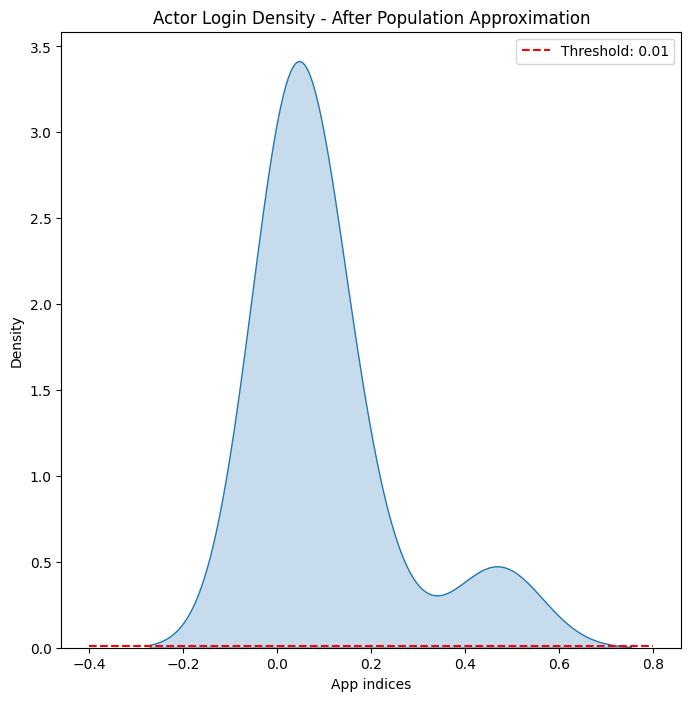}
		\caption{\label{fig:app_logins_after}Similar to the plot in fig \ref{fig:app_logins_before}, we plot the application login density for each actor. The x-axis reflects the encoded applications, the y-axis the distribution. After adjusting for the population by using the Wilson Score Confidence Interval and applying some light thresholding (in this case, we set a threshold equal to $.1$. Any app that falls under this threshold is removed.), we see a bimodal distribution emerge. What does this say about our actor? In this case, this actor accesses two sets of applications, one more frequently than the other. We can then exploit this behavior to catch for outliers} 
\end{figure}

	Utilizing this score interval, we generate a confidence interval of the true login frequency as shown in Table.\ref{table:app_confidence_interval}.  Why is this useful? We now have a confidence interval assigned to each application and each actor's usage of that application, which now reflects information about the actual true login behavior of that user. From this, we set a probability threshold, $T$, and retain only those frequencies that are above this population threshold. This yields a superset of applications associated to each actor, which we denote as $A$. This superset is a sparse matrix where each row represents an actor, and each column an application associated to that actor.  We can then plot the corrected density estimation, with this confidence interval as shown in Fig.~\ref{fig:app_logins_after}.
	
	Our last step is creating an indicator variable that compares new logins to the matrix $A$ for each actor, and computes as follows, where a 0 reflects no change in the superset, and a 1 otherwise. 
	
	\begin{equation}
	f(x)  =
	\left\{
	    \begin{array}{lr}
		0 & \text{if } I_A(x_i) = 1 \\
		 1 & \text{if } I_A(x_i) = 0 
	\end{array}
	\right\}	
	\label{eq: fxstep}
	\end{equation}
	where $I$ reflects the indicator function for the $A$ matrix, and $x_i$ represents the actor. 
	
	This feature is fed into the model as the \textbf{known\_app} or known application feature for training.

	\subsection{Data Engineering - Encoding}
	
 Okta System Log data consists of mostly categorical variables. We seek variables that exhibit some level of covariance, such as variables related to location and events. For example, columns related to country, state, zip are obviously correlated, changing country will change state, etc. Additionally, the event features are also highly correlated. Depending on the event type, other columns (such as event outcome) can change significantly.  
	
Additionally, these columns are all strings. To reference them later and perform other statistical analysis on them, it is important that we encode these variables into a numeric type. 
	
We do so using the \emph{StringIndexer} ml feature in the Spark library. This feature assigns an index to the string column in question and orders the index either by frequency or alphabetically. We can represent this mathematically as follows: 
	
	Given some column of strings of length $a$, we can set an index as follows:
	
	\begin{equation} A = (x_0, x_1, x_2...a), x \in (1,2,3..)
	\label{eq: A}
	\end{equation}
	
	Where x is a sequence of numbers, starting from 1 up to length $a$. Again, as noted earlier, how we order the strings is determined either by frequency or alphabetical order. We use alphabetical order.
	
	We utilize encodings on columns related to location, ip, event and time (weekday, event\_hour to start with)

	\subsection{Training Data - Bootstrap Sampling}
	
The system log is an event stream, which means that, for any given point of time $x_i$, we will receive a separate set of logins describing a new set of situations and contexts. It is important that, for our autoencoder to gain an accurate sense of user behavior, our training data must accurately represent the behavior of each user. To account for this variability, we utilize stratified sampling for each user, where each user represents a strata \cite{apache_spark}. 
	
How the process works is as shown in the python snippet in Listing~\ref{lst:sampling}.  We use Apache Spark to take in a set of users, creating a data set that represents a sample for each user, and then save that to our training dataset. We repeat this process N times, to build increasing confidence in our training data's representation of a specific user.  

We note that we are not particularly concerned with capturing all aspects of a user's behavior, but rather the typical; the behavior of the user as they interact with Okta's system. In fact, it is to our model's benefit if we do not fully understand all the context behind each user. By feeding the autoencoder information about known behavior, we can bias it to find more rare, anomalous events.  

\subsection{Validation - Injecting Anomalies}

As outlined above and further detailed in our testing below, we validate our model by injecting anomalies in a test data set for each user. We take a percentage of the sampled data and then place anomalies in by location or other features of interest. We will outline our approach with location.  

It is important to remember that the problem we are trying to solve is unsupervised. There is no easily obtainable validation set that represents the threats we want our model to capture. However, we can simulate some based on our own internal research.  

For location, we simulate a few scenarios, but we start with a simple one. If, within a window of time, a user logs in from two drastically distinct locations, we can create the following validation set. The first login location (or set of locations) could be from their primary geohash set, as we outlined above. The second could be from a geohash a significant distance away. We then choose a delta of distance, in miles. We start with 1000 miles (about 1609.34 km). We then feed that anomalous geohash (adjusted by that distance) to the stream of regular data and see if the model can detect the change.  

A scenario like this is important to validate a potential hijacking of employee / user credentials. If a user's credentials are hacked, and an unauthorized user is logging in / impersonating that original user (or actor), we would assume they will login from a significantly different location than the user. A model that is sensitive to this change would improve detection speed and resolution.  
	
\subsection{Training Workflow }
	
	Our training workflow(s) consist of the following components:
	
	\begin{itemize}
		\item \textbf{ETL}: We need to flatten and add additional fields to the Okta System Raw Log data. That happens in this step.
		\item \textbf{Encoding}: The sampled dataset must be converted into integer representations of categories.
		\item \textbf{Training / Sampling}: We use stratify sampling to pull sample events for each user $N$ times and construct our training set
		\item \textbf{Validation}: We create validation sets for each user. Sets for location, event\_hour and weekday to start with.
	\end{itemize}
	
	These steps constitute a \emph{workflow}. We have several training workflows, each for specific use cases. We'll outline them at a high level. These training workflows each filter for different sources of data and seek to answer slightly different questions
	
	\begin{itemize}
		\item \textbf{Policy Evaluate Sign On}: This training set contains only \textbf{policy.evaluate\_sign\_on} event types. These events fire when a user logs into Okta, so they represent an entry point for a session. They contain information such as location, application being accessed, etc
		\item \textbf{Client Information}: This training set contains only records that have a valid client object in the json record. The client object describes the client (request) that issued the HTTP okta request. These consist of a set of events that describe client object (web browser, device, etc). You can think of this dataset as containing a broader set of events than the one above. It contains more noise, but also more nuance about a user
	\end{itemize}
	
	These two workflows both construct a set of word encodings that are fed into the autoencoder. The autoencoder trains on these encodings on a regular cadence based on the loss function. This loss function helps describe the accuracy of the autoencoder. The model is trained from this data, stored, and then used against live data in the testing workflows. 
	
	\subsection{Score Workflow}
	
Our score workflows are similar in structure to that of the training data. The key differences are outlined below: 
	
	\begin{itemize}
		\item \textbf{No Sampling}: This workflow runs on current data, rather than sampled data. We still filter the data as above, but we include live events. 
		\item \textbf{Scoring}: The trained / validated model runs against this scored data, and the loss generated from this data contribute to our overall anomalies.  Accuracy in our case will be based on our security team's actual feedback on the results. 
		\item \textbf{Live Data}: The score workflows only run on live data and are sent through the trained / validated model and scored.
	\end{itemize}

	\section{Anomaly Detection}
	Anomaly detection includes techniques to identify items that are rare or differ significantly from the normal behavior observable in the majority of the data.   There are three types of anomaly detection techniques:
	
	\begin{itemize}
	\item Supervised techniques.  These techniques require annotations on the data that consists of two classes, "normal" and "abnormal" and they learn to discriminate between those two classes.
	\item Unsupervised techniques that aim at detecting anomalies by modeling the majority behavior and considering it "normal".  Then they detect the "abnormal" or fraudulent behavior by searching for examples that do not fit well to the normal behavior.
	\item Semi-supervised techniques are in between the two above cases that can learn from both unlabeled and labeled data to detect fraudulent transactions.
	\end{itemize}
	An autoencoder can be used to model the normal behavior of the data and detect outliers using the reconstruction error as an indicator and is an example of \textit{unsupervised} learning. We chose unsupervised learning due to annotations not being available and the impracticality of annotating events manually.  Additionally, for many users, a baseline behavior cannot be established due to inconsistency of work schedules and location.  For example, a user might work out of a different coffee shop every day or work from a different country a few times per year.  This makes it difficult to establish a baseline automatically based on something like the frequency of behaviors per user.  
	
	\section{Model}
	\subsection{Autoencoders}
	Autoencoders were first introduced in the 1980s to address the problem of backpropagation without a teacher by using input data as the teacher \cite{rumelhart}. They have been used for anomaly detection in cybersecurity in fields like wireless security \cite{chen} and industrial control systems \cite{nazir}. 
	
	An autoencoder is a neural network that is trained to attempt to copy its input to its output.  Internally, it has a hidden layer $h$ that describes a \textit{code} used to represent the input.  The network may be viewed as consisting of two parts:  an encoder function $h=f(x)$ and a decoder that produces a reconstruction $r=g(h)$.  If an autoencoder succeeds in simply learning to set $g(f(x))=x$ everywhere, then it is not especially useful.  Instead, autoencoders are designed to be unable to learn to copy perfectly.  Usually they are restricted in ways that allow them to copy only approximately, and to copy only input that resembles the training data.  Because the model is forced to prioritize which aspects of the input should be copied, it often learns useful properties of the data \cite{goodfellow}. 
	
	Autoencoders may be thought of as being a special case of feedforward networks and may be trained with all the same techniques, typically mini-batch gradient descent following gradients computed by back-propagation.  Unlike general feed forward neural networks, autoencoders may also be trained using recirculation, a learning algorithm based on comparing the activations of the network on the original input to the activations of the reconstructed input \cite{goodfellow}.  
	
	Copying the input to the output sounds useless at first glance, but we are typically not interested in the output of the decoder.  Instead, we hope that training the autoencoder to perform the input copying task will result in $h$ taking on useful properties \cite{goodfellow}.
	
	One way to obtain useful features from the autoencoder is to constrain $h$ to have a smaller dimension than $x$.  An autoencoder whose code dimension is less than the input dimension is called \textit{under-complete}.  Learning an under-complete representation forces the autoencoder to capture the most salient features of the training data \cite{goodfellow}.  
	
	The learning process is described simply as minimizing a loss function
	
	\begin{equation}
	L(x,g(f(x)))
	\label{eq:generic_loss}
	\end{equation}
	
	where $L$ is a loss function penalizing $g(f(x))$ for being dissimilar from $x$, such as the mean squared error.   When the decoder is linear and $L$ is the mean squared error, an under-complete autoencoder learns to span the same subspace as the more traditional machine learning approach, principal component analysis (PCA).  In this case, an autoencoder trained to perform the copying task has learned the principal subspace of the training data as a side effect \cite{goodfellow}. 
	
	Autoencoders with nonlinear encoding functions $f$ and nonlinear decoder functions $g$ can thus learn a more powerful nonlinear generalization of PCA.  Unfortunately, if the encoder and decoder are allowed too much capacity, the autoencoder can learn to perform the copying task without extracting useful information about the distribution of the data \cite{goodfellow}. 
	
\begin{figure}[h]
	\caption{Autoencoder architecture example using multi-loss between input entity embedded vector representations of 4 features.  For each feature, $Rec.i$, a loss $L_i$ between it and the original entity embedded vector, $EE_i$, is calculated and raised by a weight $w_i$.  The losses are then summed.  For the diagram example below, the number of features, $n$, is 4.}
	\centering
	\includegraphics[scale=0.40]{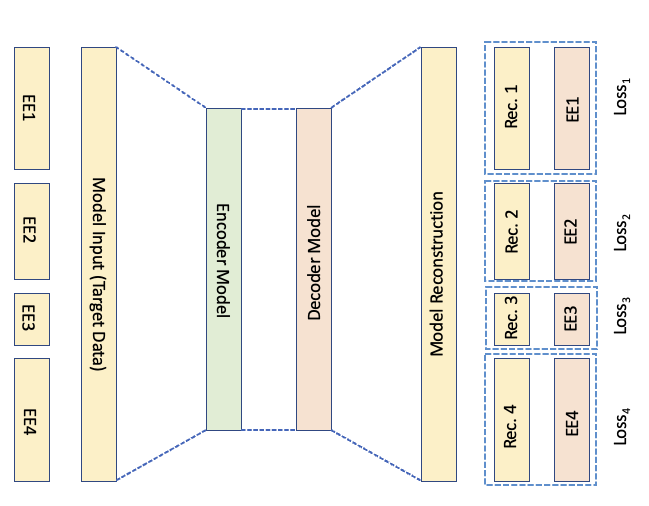}
	\label{fig:autoencoderarch}
\end{figure}

	For this project, we are using an architecture similar to Fig. ~\ref{fig:autoencoderarch}.

	\subsection{Entity Embedding}
Traditionally, the way to deal with categorical data has been one hot encoding - a method where the categorical variable is broken into as many features as the unique number of categories for that feature and for every row, a 1 is assigned for the feature representing that row's category and the rest of the features are marked 0. There are issues with one hot encoding. For categories with lots of unique features, or high cardinality, we get a lot of sparse data. Also, each vector is equidistant from every other vector which causes us to lose the value of relationships between variables. Embeddings are a solution to dealing with categorical variables with high cardinality which avoids the pitfalls of one hot encoding  \cite{cheng}.
	
	Formally, an embedding is a mapping of a categorical variable into an n-dimensional vector.  This provides us with 2 advantages.  First, we limit the number of columns we need per category.  Second, embeddings by nature intrinsically group similar variables together.  
	
	Suppose we want to use the day of the week as a feature in our autoencoder.  We create a 7 x 4 matrix mapping a day of the week to each row and initialize the tensor.  We then replace a specific day of the week with its corresponding vector as shown in Table~\ref{table: daysofweek}.

This matrix now allows us to discover non-linear relationships among variables. As opposed to one hot encoding, where a day of the week can only ever be a single value, embeddings transform the day of the week into a 4-dimensional concept. After training our model, we may discover that this table contains semantic meaning. For example, Saturday and Sunday could be more closely related than say Saturday and Wednesday. Why did we choose 4? There is no steadfast rule on how to do this, but a good heuristic is to take half the number of unique values then add one.  

For each categorical variable we initialize a random embedding matrix as m x D where m is the number of unique levels of categorical variable (ex. Monday, Tuesday…), and D is the desired dimensions for representation (ex. 4).

\begin{equation}
\begin{pmatrix} 
	w_{11}& w_{12}& ... \\ 
	...& ...& ...\\
	 w_{m1}& ... & w_{mD}\\
\end{pmatrix} 
\label{eq:pmatrix}
\end{equation}
Then for each forward pass through the neural network a lookup is done for the given level from the embedding matrix which gives a 1 x D vector. 

\begin{equation}
[w_{11}, w_{12}, ... w_{1D}]
\label{eq: prow}
\end{equation}

Next, we append this 1 x D vector to our input vector. While doing backpropagation we are updating these embedding vectors optimizing for minimizing loss. Typically, inputs are not updated but embedding matrices are a special case. 
	
	\subsection{Loss Function}
	
The loss function is very important in machine and deep learning. Let's say that you are working on any problem, and you have trained a machine learning model on the dataset and are ready to put it in front of your client. How can you be sure that this model will give the optimum result? Is there a metric or a technique that will help you quickly evaluate your model on the dataset? Here is where loss functions come into play. 
	
A loss function is a function that maps an event or values of one or more variables onto a real number intuitively representing some "cost" associated with the event. For an autoencoder, the loss function provides a metric for how similar the reconstructed input is to the original input.  
	
There are many different loss functions used for different purposes. For this work, we are using Dice Loss. Dice Loss originates from Sørensen–Dice coefficient, which is a statistic developed in 1940s to gauge the similarity between two samples. Mathematically it is represented as, 
	
	\begin{equation}
	L_d = \frac{2\sum^N_i p_i g_i}{\sum^N_i p^2_i + \sum^N_i g^2_i}
	\label{eq: loss}
	\end{equation}
	where $L_d$ is the dice coefficient, $p_i$ and $g_i$ represent elements of the reconstructed and original input vectors, respectively.  
	
	For this problem, we compute a dice coefficient for each feature-reconstructed feature pair.  For example, we would have one for day of the week, client device, client browser, etc.
	
	\section{Anomaly Detection with Autoencoders}
	Given an input feature $x_i$, and reconstructed input, $g(f(x_i))$, we compute the dice coefficient, 
	
	\begin{equation}
	D_n = L_d(x_n, g(f(x_n))), \forall n \in {1,...,n}
	\label{eq:dicen}
	\end{equation}
	
	where $n$ is the number of input features.  To compute the total loss per sample, we sum the dice loss coefficient for each feature $D = (D_1, D_2,...,D_n)$, given as 
	
	\begin{equation}
	\overline{D} = \left( \prod^n_{i=1} D*w_i \right)
	\label{eq: D}
	\end{equation}
	
	where $w = (w_1, w_2,...,w_n)$ are the weights assigned to each feature. The weights, $w$, are determined by severity of the feature if it changes.  For example, a change in country is more important than a change in city within the same state and so the weight for country could be made higher.  Once $\overline{D}$ is computed for all events in the dataset, we compute the standard deviation of $\overline{D}$ across all events,
	
	\begin{equation}
	\sigma = \sqrt{\frac{1}{N}\sum^N_{i=1}(\overline{D}_i - \mu)^2},
	\label{eq: stdev}
	\end{equation}
	
	where 
	
	\begin{equation}
	\mu = \frac{1}{N}\sum^N_{i=1}\overline{D}_i
	\label{eq: mean}
	\end{equation}
	
	Now a threshold can be set to define severity of anomalies based on $\sigma$ which is the standard deviation above the mean dice coefficient across all events.
	
	\section{Simulated Anomalies}
	
\begin{figure}[h]
	\caption{Loss at each epoch for a single actor.}
	\centering
	\includegraphics[scale=0.40]{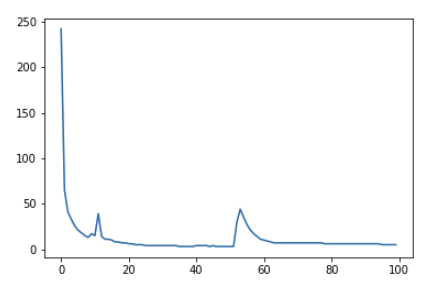}
	\label{fig:loss_epoch}
\end{figure}

We randomly chose 18 existing users. Each user had sufficient data volume. The data was split into two subsets: training and validation. For the validation set, we simulated anomalies for geohash, event hour, and day of the week. We treat these as ground truth. We train a model for each individual actor due to each having different variances in their behavior. Therefore, a loss threshold computed for one actor will not apply to another. If we computed a global threshold for all actors, it would increase our false positives. We then use each model to score the validation set with injected anomalies. An example loss for one of the actors at each epoch is shown in Fig. \ref{fig:loss_epoch}.

We compute the standard deviation of the loss between the input and output of the testing set. We compute the standard deviation of the loss between the input with simulated anomalies and its reconstruction. We use threshold $n*\sigma$ where $n = 0...10$ in increments of $0.1$. If the standard deviation between the input and the reconstructed input is above the threshold, we classify it as an anomaly and if below the threshold it is classified as normal.   

We then compute the F1 score for each $n$, 

\begin{equation}
F1 = 2\frac{P*R}{P+R}= \frac{2*TP}{2TP+FP+FN}
\label{eq: f1}
\end{equation}

The F1 score is a measure of a test's accuracy. It is calculated from the precision, P, and recall, R, of the test, where the precision is the number of true positives (TP) divided by the number of all positive results (TP+FP), and the recall is the number of TPs divided by the number of all samples that should have been positives (TP+FN). The F1 score is the harmonic mean of precision and recall, representing both in a single metric. 

	With an array of F1 scores, $F1 = (F1_0, F1_{0.1},...,F1_{10})$ for each $n$, we choose the max F1 score which in turn gives us the standard deviation and $n$ used in the threshold.  Using the F1 optimized model, standard deviation, and $n$ for each actor, we can not score future data and make a determination of their anomaly status.
	
	To do this for new incoming data, we use the mean loss for each actor from the training module, the derived $n*\sigma$ 
	for each actor, and the model for each actor. We run the new data through the autoencoder model, compute the losses between the input and reconstructed input, compute the standard deviation with respect to the mean from the training module, and apply the threshold.

	\section{Results}
	
We injected anomalies in the validation data for location, event hour, and day of the week, however, only location showed a strong correlation. We believe this is due to how we chose to featurize event hour and day of the week. For this experiment, day of the week was encoded as 1 through 7 and event hour 1 through 24. However, we believe the granularity should be reduced further so that day of the week is 1 or 0 for weekday and weekend respectively, and event hour will be 1 and 0 representing whether the log in time is in a distribution based on the actor's past behavior or not. These ideas will be explored in a future experiment. 
	
\begin{figure}[h]
	\caption{Confusion matrix for a single actor.}
	\centering
	\includegraphics[scale=0.40]{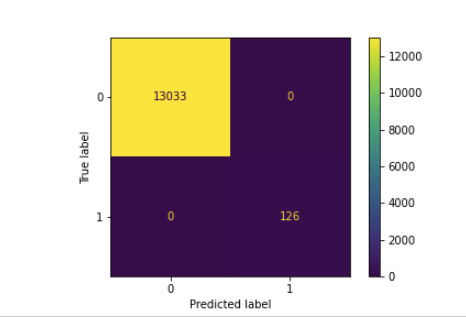}
	\label{fig:confusion_matrix}
\end{figure}

For location, we see strong signal. Fig.~\ref{fig:confusion_matrix} is an example confusion matrix for a single actor. In the validation data associated with this actor, there were 126 simulated anomalies and as shown in the confusion matrix all 126 simulated anomalies were identified without any FP or FN.  For 13 of the 18 users chosen, a near perfect F1 score was found  ($0.97 > f1 \geq 0.70$).  2 of the 18 and excellent F1 score was found  ($0.7 > f1 \geq 0.5$) and for 2 of 18 a fair F1 score was found $(\geq0.5)$.  The last 2 users were service users.  Results shown in Table~\ref{fig:results} where each row is for a particular actor.

\section{Conclusions and Next Steps}
	
	Using an autoencoder, we show that location anomalies in OKTA logs can be identified with high precision.  We did not find strong signal for event hour and day of the week, but as discussed above we hope to find strong signal by reducing granularity in these features.  We also plan to experiment with anomalies in other available features that are used for training such as operating system, device type, etc.  We are in the process of scaling out the production workflow and will test on 100 users next.  The model that is trained will be registered in MLFLOW if the F1 score is high, and will not be retrained for a given actor.  If the F1 score is poor, the model will be retrained whenever a new login attempt occurs.  
	
	Of additional interest is the scenario of impossible travel.  Meaning, a location change has been detected for a given actor, and not enough time has passed to reasonably travel from that actor's previous location.  This anomalous travel might be detectable using time series based machine learning methods such as an LSTM autoencoder or ARIMA.

\section{Acknowledgments}
The authors thank the anonymous reviewers for their valuable suggestions.  The authors would also like to thank Hemanth Adusumalii for his invaluable insights on Okta data.

\begin{itemize}
		\item[]
			\begin{figure}[h]
			\centering
			\includegraphics[width=0.05\textwidth]{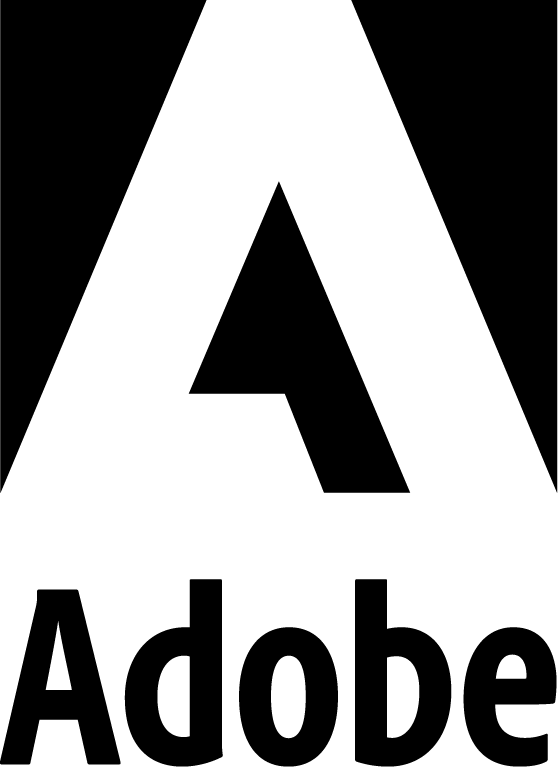}
		\end{figure}
		\item[] \fontsize{8}{10}\emph{Copyright and Legal Disclaimer: \textcopyright  2023 Adobe. All rights reserved. Adobe and the Adobe logo are either registered trademarks or trademarks of Adobe in the United States and/or other countries. These materials are intended for informational purposes only. Provision of this information does not entitle the recipient to any contractual or other rights. Furthermore, while efforts have been made to assure the accuracy of the information as of the date it has been provided, no representation is made that such information is accurate and complete, and Adobe undertakes no obligation to update this information as Adobe's products or supporting infrastructure change.}
	\end{itemize}

\bibliographystyle{unsrt}
\bibliography{reference.bib}

\clearpage

\begin{minipage}{\linewidth}
	\begin{lstlisting}[style=pythonstyling, caption={Sampling strategy. This python code shows  a simple function that performs stratify sampling on a dataset, given a set of actors X. The samples are created for each actor and then merged into a final dataset, which can be used for training.}, captionpos=b, label={lst:sampling}]
	def stratify_sampling(dataset, X, output_table):
	    fractions = {x:.1 for x in X}
	    N = 10
	    for _ in range(N):
	        for user in X:
		    sample = dataset.sampleBy('actor_id', fractions=fractions, seed=1)
		    sample.write.mode('append').saveAsTable(output_table)
	\end{lstlisting}
\end{minipage}

\begin{table}[]
\begin{tabular}{lllll}
app. instance counts&  actor sum&  WS-CI&  max(WS-CI)&  mean(WS-CI)\\ \hline
 1&  4&  [0.078, 0.567]&  0.567& 0.323 \\
 2&  4&  [0.230, 0.770]&  0.770&  0.500\\
 1&  4&  [0.078, 0.567]&  0.567&  0.323\\
 2&  8&  [0.109, 0.476]&  0.476&  0.293\\
 4&  8&  [0.294, 0.706]&  0.706& 0.500\\ \hline
\end{tabular}
\caption{This table shows the Wilson Score Confidence Intervals for a set of application login frequencies by actor. Each row represents an application login grouping for  a single actor. In this case, the first three rows represent logins for an actor $A$, and the last two represent the logins for a second actor $B$. The first column, \emph{app. instance counts} reflects the frequency of times a user logged in that particular app. \emph{actor sum} represents the total number of logins by the actor. \emph{WS-CI} reflects the confidence interval captured for that application login (uses the 1st and 2nd column to compute). The last two columns \emph{max(WS-CI) and mean(WS-CI)} calculate the max and mean respectively of this Wilson Score Confidence Interval. The mean is what we use as the \emph{typical} login probability for that actor and application, as it tends to pull down login probabilities to more reasonable estimates. }
\label{table:app_confidence_interval}
\end{table}

\begin{table}[]
	\begin{tabular}{lclclclcl} \hline
 	Sunday&0.4  &-0.3  &0.6  & 0.1 \\
 	Monday&0.2 &0.2  &0.5  &-0.3  \\
 	Tuesday&0.1  &-1.0  &1.3  &0.9  \\
 	Wednesday&-0.6  &0.5  &1.2  &0.7  \\
 	Thursday&0.9  &0.2  &-0.1  &0.6  \\
 	Friday&0.4  &1.1  &0.3  &-1.5  \\
	Saturday&0.3  &-0.2  &0.6  &0.0 \\ \hline
	\end{tabular}
	\caption{Example of entity embeddings for days of the week.}
	\label{table:daysofweek}
\end{table}

\clearpage
\begin{table}[]
\begin{tabular}{llllll}\hline
 f1 score&number of samples  &precision  &recall  &standard deviation  &std. dev. coefficient  \\ \hline
 1.0&  1271&  1.0&  1.0&  0.136& 3.3 \\
 1.0&  1061&  1.0&  1.0&  0.314&  2.5\\
 1.0&  1315&  1.0&  1.0&  0.568&  1.4\\
 1.0&  1264&  1.0&  1.0&  0.169&  5.6\\
 1.0&  398&  1.0&  1.0&  0.661&  2.4\\
 1.0&  506&  1.0&  1.0&  0.448&  2.0\\
 1.0&  244&  1.0&  1.0&  0.242&  6.7\\
 1.0&  354&  1.0&  1.0&  0.231&  4.1\\
 1.0&  715&  1.0&  1.0&  0.195&  5.0\\
 1.0&  678&  1.0&  1.0&  0.175&  6.3\\
 1.0&  13159&  1.0&  1.0&  0.250&  2.2\\
 0.996&  13414&  1.0&  0.993&  0.230& 8.5 \\
 0.984&  14164&  0.970&  1.0&  0.162&  0.4\\
 0.920&  65614&  0.852&  1.0&  0.274&  3.4\\
 0.727&  1169&  0.571&  1.0&  1.00&  0.4\\
 0.588&  710&  0.417&  1.0&  0.211&  4.7\\
 0.060&  43195&  0.031&  1.0&  2.68&  0.1\\
 0.022&  1144&  0.011&  1.0&  1.18& 0.9\\ \hline
\end{tabular}
\caption{Results table showing various metrics.  Each row represents 1 of 18 actors.  For 13 of the 18 actors, a near perfect F1 score was found ($\geq 0.97$).  2 of the 18 had excellent F1 scores ($0.97 > f1 \geq 0.70$).  For 2 of 18 a fair F1 score was found ($0.7 > f1 \geq 0.5$).  The last 2 users were service users.}
\label{fig:results}
\end{table}

\clearpage
\clearpage
\clearpage
\clearpage
\clearpage

\clearpage
\clearpage	
	
\clearpage	
	
\clearpage

\end{document}